\pdfoutput=1
\documentclass[10pt, a4paper]{article}

\usepackage{lrec-coling2024} 

\usepackage{natbib}
\usepackage{multibib}
\makeatletter
\def\@mb@citenamelist{cite,citep,citet,citealp,citealt,citepalias,citetalias}
\makeatother
\newcites{languageresource}{~}

\usepackage{graphicx}
\usepackage{tabularx}
\usepackage{soul}
\usepackage{comment}

\usepackage{booktabs}
\usepackage{color}
\usepackage{comment}
\usepackage{numprint}
\usepackage{hyperref}

\usepackage{xcolor}
\usepackage{hyperref}
 \definecolor{darkblue}{rgb}{0, 0, 0.5}
  \hypersetup{colorlinks=true, citecolor=darkblue, linkcolor=darkblue, urlcolor=darkblue}

\usepackage{xstring}
\usepackage{xspace}

\usepackage{color}

\newcommand{\yoruba}{Yor\`{u}b\'{a}\xspace}

\newcommand{\iroyin}{\`{I}r\`{o}y\`{i}nSpeech}

\newcommand*{\corpusa}{\textsc{Partition A}\xspace}
\newcommand*{\corpusb}{\textsc{Partition B}\xspace}
\def\thickhline{\noalign{\hrule height.8pt}}

\title{\`{I}r\`{o}y\`{i}nSpeech: A multi-purpose Yor\`{u}b\'{a} Speech Corpus}

\name{Tol\'{u}l\d{o}p\d{\'{e}} \'{O}g\'{u}nr\d{\`{e}}m\'{i} $^{1}$, K\d{\'{o}}l\'{a} T\'{u}b\d{\`{o}}s\'{u}n$^{2}$,  Anuoluwapo Aremu$^{2}$, Iroro Orife$^{3}$, \\ \large{\textbf{David Ifeoluwa Adelani$^{4}$}}} 
\address{$^{1}$Stanford University, $^{2}$\yoruba Names, $^{3}$Niger-Volta LTI, $^{4}$University College London \\
tolulope@cs.stanford.edu, project@yorubaname.com, iroro@alumni.cmu.edu, d.adelani@ucl.ac.uk \\ }

\abstract{
We introduce \iroyin, a new corpus influenced by the desire to increase the amount of high quality, contemporary \yoruba speech data, which can be used for both Text-to-Speech (TTS) and Automatic Speech Recognition (ASR) tasks. We curated about \numprint{23000} text sentences from news and creative writing domains with the open license CC-BY-4.0.
To encourage a participatory approach to data creation, we provide \numprint{5000} curated sentences to the Mozilla Common Voice platform to crowd-source the recording and validation of \yoruba speech data. In total, we created about 42 hours of speech data recorded by 80 volunteers in-house, and 6 hours of validated recordings on Mozilla Common Voice platform. Our TTS evaluation suggests that a high-fidelity, general domain, single-speaker \yoruba voice is possible with as little as 5 hours of speech. 
Similarly, for ASR we obtained a baseline word error rate (WER) of 23.8.\\ 
\newline  }

\begin{document}

\maketitleabstract

\section{Introduction}
\label{sec:intro}
Speakers of many African languages have no access to voice-enabled applications in their native languages. One reason is that these technologies all require models for speech recognition and speech synthesis, trained on large datasets of high-fidelity speech and text~\citep{Ritchie2022LargeVS,meyer2022bibletts}. 

To address this challenge, there have been several efforts to build large-scale multilingual datasets and models by automatically aligning speech and text~\citep{pmlr-v202-radford23a,Zhang2023GoogleUS,pratap2023mms}. The results however, are often of poor quality for low-resource African languages due to the dearth of high-quality audio-text pairs and unsatisfactory out-of-domain generalization. Other efforts have focused on building benchmark datasets for over 100 languages using high-quality but small-scale training data~\citep{fleurs, shi23g_interspeech,Shi2023FindingsOT}. 

In this paper, we focus on \yoruba, a West African language with over 40 million L1 speakers, yet under-represented in contemporary speech research.  
There have been a 
number of efforts to build datasets for \yoruba speech tasks \cite{Odejobi2004ACM, ajadi2007quantitative, Akinwonmi2013APT, Afolabi2014DevelopmentOT, Dagba2016DesignOA,Niekerk2012ToneRI, van2015lagos,gutkin2020developing}. The datasets are either too small to do speech processing effectively or are single speaker, single domain, as is the case for BibleTTS \cite{meyer2022bibletts}. Our corpus extends the scope of previous work to address multiple speech application domains.

We introduce the \iroyin --- a new dataset created to increase the amount of high quality, contemporary \yoruba speech. 
The dataset has a total of 42 hours of audio, recorded by 80 volunteers. 
We curated text sentences from the news and creative writing domains under an open license, CC-BY-4.0. We also provide \numprint{5000} sentences to the Common Voice \cite{commonvoice} platform to crowdsource voice recordings online\footnote{\url{https://commonvoice.mozilla.org/yo}}. We provide extensive baseline experiments using state-of-the-art approaches for TTS and ASR. The code and data will be made freely available on \url{https://github.com/Niger-Volta-LTI/yoruba-voice}.

\section{The Yor\`{u}b\'{a} Language}
 The \yoruba  language is native to south-western Nigeria, Republic of Benin, and Republic of Togo. It is one of the national languages of Nigeria also spoken in other countries like Ghana, Côte d'Ivoire, Sierra Leone, Cuba and Brazil. The language belongs to the Niger-Congo family in the Volta-Niger sub-group, and is spoken by over 40 million native speakers~\cite{Ethnologue2019}, making it one of the most widely spoken African languages. 
 
 \yoruba has 25 letters without the Latin characters  (c, q, v, x and z) and with additional characters ({\d e}, gb, {\d s}, {\d o}). 
 There are 18 consonants, 
 seven oral vowels  (a, e, {\d e}, i, o, {\d o}, u), five nasal vowels, 
 (an, {\d e}n, in, 
 {\d o}n, un) and syllabic nasals 
({\`{m}}, {\'{m}}, {\`{n}}, {\'{n}}). \yoruba is a tonal language with three tones: low, middle and high. These tones are represented by the grave (``\textbackslash
''), optional macron (``$-$'') and acute (``/'') accents respectively.  These tones are applied on vowels and syllabic nasals, but the mid tone is usually ignored in standard \yoruba orthography. These \textit{diacritics} are important for correct pronunciation and lexical disambiguation. 

\section{The \`{I}r\`{o}y\`{i}nSpeech Corpus}

\label{sec:corp-creation}
\subsection{Preparation of text sentences}
In contrast to other \yoruba datasets based on Biblical or religious texts, our goal was to combine news data and fictional texts to create a modern, multi-purpose speech dataset \cite{gutkin_49562,meyer22c_interspeech}. The corpus text was obtained from two sources, firstly the MENYO-20k dataset \cite{adelani-etal-2021-effect}, an open-source, multi-domain English-\yoruba machine translation corpus and secondly, the \yoruba portion of the MasakhaNER 2.0 dataset~\cite{adelani-etal-2022-masakhaner} (i.e MasakhaNER-YOR) based on the Asejere newspaper\footnote{\url{https://www.asejere.net/}}. The primary source of the MENYO-20k dataset is the Voice of Nigeria newspaper\footnote{\url{https://yoruba.von.gov.ng/}}, published by the Nigerian government. 
We restrict our selection of corpus text to the above published datasets for two reasons (1) they have a non-restrictive license, and  (2) the \yoruba sentences have been further verified for quality issues, for example missing diacritics in the original crawled Asejere and Voice of Nigeria articles. Overall, we obtained \numprint{3048} sentences from Voice of Nigeria, \numprint{2932} sentences from Global Voices, and \numprint{5135} sentences from Asejere. In total, this gives us \numprint{11115} sentences.

In order to obtain more sentences to reach our goal of 40 hours of speech, we added sentences extracted and modified from unpublished short stories previously translated into \yoruba by the second author. 
These texts were selected to broaden the domain of the vocabulary used in the dataset. In addition, we split-up long sentences
and asked volunteers to manually generate new sentences with similar themes or context as the original seed sentences. They also cross-checked each sentence for errors. In total, we had to manually generate about \numprint{12000} sentences.
We then cleaned up the data to create a final script. To ensure the sentences were of high-quality, we verified that diacritics were properly applied on each word and revised offensive or divisive religious terms within the text to reflect a neutral tone. Next, we modified the text for clarity and length to facilitate pronunciation, and localized non-Yor\`{u}b\'{a} words into \yoruba. We list below a few examples of the types of names and places which were localized: Kaduna to \d{\`{O}}y\d{\'{o}}, Zamfara to O\`{n}d\'{o}, United States to \`{I}l\'{u} \d{O}ba, Buhari to B\`{u}h\'{a}r\'{i} and Kenya to K\d{\'{e}}\'{n}y\`{a}.

\subsection{Recording of text sentences}

\begin{table*}[t]
    \centering
    \scalebox{0.93}{
    \begin{tabular}{crrrr}
    \toprule
     & \textbf{\# hours} & \# \textbf{utterances} & \textbf{Corpus partitions used} & \textbf{\# Unique sentences used} \\
    \toprule
     In-house ASR  & 26h 00m & \numprint{20000} & \corpusa & \numprint{20000}  \\
     Common Voice ASR  & 6h 00m & \numprint{5000} & \corpusa & - \\
     In-house TTS & 10h 11m & \numprint{9000} &  \corpusa \& \corpusb &  \numprint{3000} \\

     \midrule
    Total & 42h 11m & \numprint{34000} & - & \numprint{23000} \\
    \bottomrule
    \end{tabular}
    }
    \caption{A summary of dataset statistics. Some of the utterances used for TTS recording (i.e. 6K utterances) and for Common Voice (5K utterances) are subsets of the \corpusa. }
    
    \label{tab:dataset_stats}
\end{table*}

\subsubsection{Corpus partitions}
\autoref{tab:dataset_stats} provides the details of the recorded utterances and text preparation. Our text preparation yielded a total of 
\numprint{23000} sentences which were used to record audio of both ASR and TTS. 
Our initial setup divided the corpus into two parts: (1) \textbf{\corpusa}, contains \numprint{20000} sentences, primarily for the recording of ASR audio and (2) \textbf{\corpusb}, contains \numprint{3000} reserved for TTS recordings. 

\paragraph{ASR recording} We recorded in-house all \numprint{20000} sentences in \corpusa, or some 26 hours for ASR. 
They were recorded by 80 different volunteers, each recording 250 lines during one-hour studio sessions. Additionally, we added \numprint{5000} sentences to the Mozilla Common Voice crowd-sourcing platform, which were recorded by 108 volunteers on the Common Voice website, yielding 6 hours of speech. The sentences used to record on Common Voice were selected from the sentences in \corpusa. We note for completeness, that the ASR experiments in this study did not \textit{yet} make use of these 6 hours. We hope that by setting up Common Voice for \yoruba, native speakers everywhere will be encouraged contribute. 

\paragraph{TTS recording} We recorded all \numprint{3000} sentences in \corpusb by two speakers, one male and one female, ages 25 to 30. This resulted in 3 hours 35 minutes of data, which was below our goal of 10 hours. Since there were no additional curated sentences, we decided to record supplementary sentences obtained from \corpusa to reach our 10 hour goal. The final TTS corpus contains \numprint{9000} sentences yielding 10 hours 11 minutes of speech. 


\smallskip
\smallskip
All volunteers, speakers of standard North West \yoruba\footnote{\url{http://www.africa.uga.edu/Yoruba/yorubabout.html}}, were screened for dialect uniformity, and ranged in age from 18 to 69 years. The \numprint{9000} lines for the single speaker (TTS) partition had one male and one female volunteer, while the \numprint{20000} lines multi-speaker (ASR) partition had 80 volunteers, 37 male and 43 female. The studio volunteers were each provided with a token gift, as a gesture of appreciation of their time and efforts recording.

\subsubsection{Recording}
To create an acoustically suitable environment for recording, we obtained a portable vocal booth. The recording equipment comprised an AT 2020+ USB microphone, USB cables, and a 2022 M1-Series Macbook Pro.

The first five hours of audio were recorded with Audacity, a free, open-source digital audio editor and recording application. To divide each of these hour-long recordings into a short file for each sentence required many more additional hours of manual post-editing work. To solve this problem, the team developed a custom application for creating speech corpora, dubbed \textit{\yoruba Voice SpeechRecorder}\footnote{\url{https://github.com/Niger-Volta-LTI/yoruba-voice-speech-recorder}} \cite{yvspeechrecorder}.

The app works by reading a prepared text file, usually with 250 sentences and displays each line of text to be read in order. It also provides transport controls to enable recording, playback and file-management or deletion, in the case of multiple takes. Finally, the tool saves individual audio files to the hard-drive, for each sentence, and updates a metadata index, which can be used to programmatically prepare training examples. Over 65\% of the total lines recorded in-house were recorded using the SpeechRecorder app.


\subsubsection{Post-production}
We had four forms of post-processing. Where possible, recordings that had issues which could be manually fixed, were repaired by removing simple audio artifacts and speech disfluencies. In situations where the recording did not correspond with the text but the utterance remained grammatical, we did not rerecord the utterance but rather edited the text sentence to match the audio. 

We also fixed tone marking, spelling, or semantic mismatches. Words like ``n\'{i} il\'{e}'' (into the house) or ``s\'{i} ib\d{\`{e}}'' (to there) are often contracted to ``n\'{i}l\'{e}'' and ``s\'{i}b\d{\`{e}}'' respectively in spoken \yoruba and were amended in the text sentence accordingly. 

If the audio files had any issues which rendered them unusable, then we re-recorded, usually with a different volunteer of the same gender, introducing thusly, a new, different speaker ID. Some of the issues encountered include: (1) \textit{Disfluencies}: hesitations, stammers, clicks, sniffs, etc. (2) \textit{External noises}: paper rustling, microphone touching, intrusive voices, electronic notification beeps, etc (3) \textit{Audio fidelity}: low or uneven audio levels, clipping or distortion, or otherwise unintelligible words (4) 
\textit{Incorrect dictation} which could not be fixed by changing the script. 

\section{Speech Synthesis Experiments}
We train speech synthesis models using the single-speaker TTS partition of our dataset, resulting in both male and female voices. The \numprint{9000} utterances described in \autoref{tab:dataset_stats}, are split evenly between the male and female speakers, employing \numprint{4500} utterances for each. We train and evaluate three variants of the VITS (Variational Inference with adversarial learning for end-to-end Text-to-Speech) model\footnote{But also release Tacotron2-DCA and Glow TTS models trained with the same data} as follows: 
\begin{itemize}
  \item Domain adaptation from existing BibleTTS \yoruba checkpoints~\citep{meyer2022bibletts}.
  \item Training a VITS model end-to-end from scratch with 5 hours of data  
  \item Training VITS models from scratch without diacritics in the training text
  \item Training a VITS model end-to-end from scratch with different data scales i.e. number of utterances: \numprint{100}, \numprint{500}, \numprint{1000}, \numprint{2000} and \numprint{4500}.
\end{itemize}

\paragraph{Model training} We trained the VITS models using the Coqui TTS toolkit \citep{meyer2022bibletts}. We use the AdamW optimizer~\citep{loshchilov2018decoupled}  with betas \{0.8, 0.99\}, weight decay of 0.01, an initial learning rate of 0.0002 decaying exponentially by a gamma of 0.999875. The models were trained with a batch size of 16 using an NVIDIA A10 GPU with 24GB of GPU memory. For the domain adaptation, we fine-tuned for 100K iterations steps, while when training from scratch, 500K iterations steps were required. Finally, for the last experiments, all models were trained for 100K iterations steps since the model performance had typically started to converge. 

\paragraph{Model evaluation} For subjective evaluation, we ran Mean Opinion Score (MOS) and MUSHRA tests \citep{2014MethodFT} via an online web application, with 70 participants. We also did objective evaluation of the models' output, measuring the mean Mel Cepstral Distortion~\citep{kominek08_sltu} of selected utterances for each model. Our evaluation data is also based on the \textit{news domain} but on samples \textit{unseen} during training. 

\subsection{Does training with diacritics affect synthesised voices?}

The importance of tone, represented orthographically by diacritics in \yoruba led us to question whether there would be a difference between the speech produced by models trained with or without diacritics. For the male and female voices, we train TTS models from scratch with and without diacritics, presenting the results in \autoref{tab:tts-diac}.

\begin{table}[!h]
    \centering
    \resizebox{\columnwidth}{!}{%
    \begin{tabular}{lccc}
        \thickhline
        Model & MOS$\uparrow$ & MUSHRA$\uparrow$ & MCD$\downarrow$ \\
        \thickhline
       With diacritics, Male  & \textbf{3.98 }& \textbf{60.91}& \textbf{7.47} \\
       No diacritics, Male   & 1.86 & 18.97 & 8.93 \\
         \hline
      With diacritics, Female   & \textbf{2.82} &\textbf{48.24} & \textbf{6.26} \\
      No diacritics, Female  & 1.50 & 19.02& 11.88\\
         \thickhline
    \end{tabular}
    }
    \caption{Results of experiments training TTS models with and without diacritics. The Mean Opinion Scores (MOS), MUSHRA scores and Mel-Cepstral Distortion (MCD) is measured. Across the board, models trained with diacritics perform better than those not trained with diacritics.}
    \label{tab:tts-diac}
\end{table}

Across voices and evaluation metrics, we find that training models with diacritics leads to more natural sounding speech. The female voice does not have as high a MOS or MUSHRA score as the male voice, indicating a less natural sounding voice in comparison.

\subsection{Does continued pre-training result in a more natural voice?}

The availability of BibleTTS models \cite{meyer2022bibletts} in \yoruba provided the opportunity to use our dataset to continue pre-training TTS models which already produce natural sounding speech. The BibleTTS \yoruba voice uses a single male voice. We wanted to observe whether starting from a trained TTS model would lead to a more natural sounding voice than using our data alone. To test this hypothesis, we trained models from scratch and continued pre-training the BibleTTS \yoruba checkpoint. Results of the model evaluation are in Table \ref{tab:tts-bible}.

\begin{table}[!htbp]
    \centering
    \resizebox{\columnwidth}{!}{%
    \begin{tabular}{cccc}
        \thickhline
        Model & MOS$\uparrow$ & MUSHRA$\uparrow$ & MCD$\downarrow$\\
        \thickhline
       Male  & 3.98 & \textbf{60.91}& \textbf{7.47}\\
       BibleTTS $\rightarrow$ Male   & \textbf{4.22} & 52.59 & 8.86 \\
         \hline
      Female   & \textbf{2.82} &\textbf{48.24} & \textbf{6.26}\\
      BibleTTS $\rightarrow$ Female   & 2.57 & 41.63& 8.93 \\
         \thickhline
    \end{tabular}
    }
    \caption{Subjective \{MOS, MUSHRA\} and Objective \{MCD\} evaluation results of voices trained from scratch (Male, Female) and continued pre-training (BibleTTS $\rightarrow$ Male, BibleTTS $\rightarrow$ Female).} 
    \label{tab:tts-bible}
\end{table}

For the female voice, we see that training from scratch leads to the best performance across objective and subjective metrics. This is likely due to the BibleTTS voice being male, and thus a more difficult adaptation during continued pre-training. For the male voice, the results are mixed. Although the MOS score is higher for the continued pre-training voice, the MUSHRA and MCD scores are higher for the model where we train from scratch. This means that when compared to the continued pre-training model, the from-scratch model is rated more natural. 

\subsection{How much data is required to train a synthetic voice?}

Given the low-resource setting we work in, we experiment how far we the push the limit of a few resources. We test how many utterances are required to train a model that produces natural speech. Due to the number of models already trained, we measure this solely through objective evaluation with MCD.

\begin{table}[!h]
    \centering
    \begin{tabular}{c|c}
    \thickhline
       Number of utts.  & MCD$\downarrow$ \\
       \thickhline
       100 utterances  & 7.49\\
       500 utterances  & 6.99\\
       1000 utterances  & 7.11\\
       2000 utterances  & 7.09\\
       4500 utterances  & \textbf{6.85} \\
       \hline
    \end{tabular}
    \caption{MCD of TTS models trained with varying numbers of utterances, ranging from \numprint{100} to \numprint{4500}. The model trained with the most utterances has the best performance.}
    \label{tab:mcd-fem}
\end{table}

The results for this experiment are in Table \ref{tab:mcd-fem}. The Mel Cepstral Distortion is highest for the model trained with the fewest utterances (7.49) and lowest for the model trained with the most utterances (6.85). Although the MCD does not decrease monotonically as the number of utterances increases, there is evidence that more data is better. Overall, based solely on this objective evaluation, one may reason that only 500 utterances are necessary to train a satisfactory \yoruba VITS model.

\section{Automatic Speech Recognition Experiments}
To evaluate our corpus for speech recognition tasks, we train several baseline ASR models, with the following \textbf{data split of \numprint{15000} / \numprint{1000} / \numprint{4000}} for training, development and test respectively. We explore training a conformer model end-to-end and finetuning self-supervised speech representations. The results of these experiments are in Table \ref{tab:my_label}.

\subsection{End-to-End Conformer model}
We use ESPNet to train a 12-layer Conformer model end-to-end with an RNN language model (LM) for decoding. We use unigram tokenization and have a perplexity of 54.0 on the held-out validation set.

\subsection{Finetuning wav2vec 2.0}
We finetune wav2vec 2.0 XLSR-300m~\citep{babu22_interspeech} and train an end-to-end Conformer model \citep{conformer}. For wav2vec 2.0 XLSR-300m, a massive multilingual model, pretrained for speech tasks on 128 languages, we finetune for 20 000 steps, equating to 10.67 epochs.

\begin{table}[!h]
    \centering
    \begin{tabular}{c|c}
    \thickhline
      Model   & WER$\downarrow$ \\
      \thickhline
       Conformer + RNN LM & 69.7 \\
        wav2vec 2.0 finetuned & 40.6 \\
          +bigram model & 27.6 \\
          +trigram model  & \textbf{23.8}\\
            \hline
    \end{tabular}
    \caption{Word-error-rate (WER) for end-to-end Conformer model and finetuned wav2vec 2.0. Finertuning wav2vec 2.0 outperforms training an end-to-end Conformer model from scratch.}
    \label{tab:my_label}
\end{table}

We observe that finetuning wav2vec 2.0 leads to significantly better performance versus training the Conformer model end-to-end. The addition of an $n$-gram language model further lowers the error-rate, with the trigram LM model prevailing. 

Overall, a more substantial evaluation, beyond these initial baselines, will be required to better understand the benefits of finetuning a multilingual model versus training a simpler monolingual model. Finally, we hope that these initial results will encourage the inclusion of the \yoruba language in more multilingual ASR evaluation benchmarks.

\section{Conclusion}
In this work we present an open dataset of 42 hours of high quality \yoruba speech data to be used for both Speech Synthesis and Automatic Speech Recognition research. For TTS, we remark that models trained with diactrics generate speech that is perceived as more natural than those trained without diacritcs, while models continually trained from existing models may not always sound more natural than those trained from scratch. In ASR, we see that finetuning wav2vec 2.0 with a trigram model leads to the lowest word error rate. This data will be made freely available in the hope that it will invigorate speech research and accelerate the development of technology for the \yoruba language.

\section{Ethics Statement}
For this project, we obtained the consent of all the volunteers that contributed their voice recording to the \iroyin{} project. Also, our recording does not include private or sensitive conversations that can violate our volunteers privacy since the utterances are mostly from the news domain. 

\section{Acknowledgements}
We would like to thank the reviewers for their comments and suggestions. This work was carried out with support from Imminent Translated, whose 2022 funding helped support the project. Special thanks to Mr. Bode Adedeji and Miss Aguobi Nkechinyere Faith of the Department of Linguistics, University of Lagos, for permission to place our booth in their shared office while we did some of our recordings. David Adelani acknowledges the support of DeepMind Academic Fellowship programme. This research was funded in part by a Stanford School of Engineering Fellowship to TO. Finally, we are grateful to Professor Christopher Manning and Professor Dan Jurafsky for their useful feedback on the draft.


\nocite{*}
\section{Bibliographical References}\label{sec:reference}

\bibliographystyle{lrec-coling2024-natbib}
\bibliography{lrec-coling2024-example}


\end{document}